\documentclass[lettersize, journal]{IEEEtran}
\usepackage{amsmath,amsfonts}
\usepackage{algorithmic}
\usepackage{algorithm}
\usepackage{array}
\usepackage[caption=false,font=normalsize,labelfont=sf,textfont=sf]{subfig}
\usepackage{textcomp}
\usepackage{stfloats}
\usepackage{url}
\usepackage{verbatim}
\usepackage{graphicx}
\usepackage{cite}
\usepackage{multirow}
\usepackage{tabularx}

\usepackage{tikz,xcolor}
\usepackage[hidelinks]{hyperref}
	  
\definecolor{lime}{HTML}{A6CE39}
\DeclareRobustCommand{\orcidicon}{
	\begin{tikzpicture}
	\draw[lime, fill=lime] (0,0) 
	circle [radius=0.16] 
	node[white] {{\fontfamily{qag}\selectfont \tiny ID}};
	\draw[white, fill=white] (-0.0625,0.095) 
	circle [radius=0.007];
	\end{tikzpicture}
	\hspace{-2mm}
}

\foreach \x in {A, ..., Z}{
	\expandafter\xdef\csname orcid\x\endcsname{\noexpand\href{https://orcid.org/\csname orcidauthor\x\endcsname}{\noexpand\orcidicon}}
}

\begin{document}
	\title{Deep Learning-based Filtering for Video Coding: \\
		A Survey on Architectures, Algorithms, and Complexity Analysis}
	
	\author{Young-Woon Lee\orcidA{} and
		Byung-Gyu Kim\orcidB{},~\IEEEmembership{Senior Member,~IEEE}
	\thanks{Received 26 May 2026; accepted 3 July 2026. \textit{(Corresponding author: Byung-Gyu Kim)}}
	\thanks{Young-Woon Lee is with the Department of Computer and Electronics Convergence Engineering, Sunmoon University, Asan 31460, South Korea (e-mail: yw.lee@ivpl.sm.ac.kr)}
	\thanks{Byung-Gyu Kim is with the Division of Artificial Intelligence Engineering/ICT Convergence Research Institute, Sookmyung Women’s University, Seoul 04310, South Korea (e-mail: bg.kim@sookmyung.ac.kr).}
	\thanks{Digital Object Identifier 10.1109/TCE.2026.3711657}}
	
	\markboth{IEEE Transactions on Consumer Electronics, July~2026}
	{Lee \MakeLowercase{\textit{et al.}}: Deep Learning-based Filtering for Video Coding: A Survey on Architectures, Algorithms, and Complexity Analysis}
	
	\IEEEpubid{\parbox{\textwidth}{\centering \vspace{15px} 1558-4127 © 2026 IEEE. All rights reserved, including rights for text and data mining, and training of artificial intelligence and\\
		similar technologies. Personal use is permitted, but republication/redistribution requires IEEE permission.\\
		See https://www.ieee.org/publications/rights/index.html for more information.}}
	
	\maketitle

	\begin{abstract}
		As Ultra-High-Definition (UHD) displays and immersive media services become ubiquitous in the Internet of Things (IoT) and Consumer Electronics (CE) sectors, including 8K display and mobile devices, the demand for high-efficiency video coding is unprecedented. 
		While Deep Learning-based Filtering (DLF) has emerged as a promising solution to mitigate compression artifacts inherent in standards like High Efficiency Video Coding (HEVC/H.265) and Versatile Video Coding (VVC/H.266), its deployment in CE devices is severely constrained by computational complexity, memory bandwidth, and power consumption. 
		To bridge the gap between academic research and practical deployment, this paper presents a comprehensive, hardware-oriented survey of DLF techniques. 
		We propose a systematic three-dimensional taxonomy classifying methods into (1) Integration Scheme within the Video Coding, (2) Coding Information Utilization, and (3) Network Design Strategy. 
		Unlike prior reviews, this work critically analyzes the trade-offs between Rate-Distortion (RD) performance and hardware feasibility, highlighting the evolution from heavy, performance-oriented models to lightweight, hardware-friendly architectures targeting Neural Processing Units (NPUs).
		Furthermore, we incorporate the latest standardization activities from the Joint Video Experts Team (JVET) on Neural Network-based Video Coding (NNVC) to provide realistic guidelines. 
		We also identify open challenges such as real-time inference latency and error propagation, providing a roadmap toward robust, low-power intelligent video coding in next-generation CE vision endpoints.
	\end{abstract}
	
	
	\section{Introduction}\label{introduction}
		Video services such as 4K/8K UHD, immersive Virtual Reality (VR), mobile streaming, and interactive applications continue to grow rapidly, resulting in unprecedented demand for high-quality video at low bitrates. 
		To address this need, modern video coding standards including HEVC/H.265 \cite{Sullivan2012HEVC} and the more recent VVC/H.266 \cite{Bross2021VVC} have significantly improved compression performance through advanced prediction, transform, quantization, and entropy coding tools. 
		However, because these standards rely on block-based hybrid coding, quantization inevitably introduces visual artifacts which degrade the perceptual Quality of Experience.
		
\begin{figure}[!t]
	\centering
	\renewcommand{\arraystretch}{0.8}
	\includegraphics[width=\columnwidth]{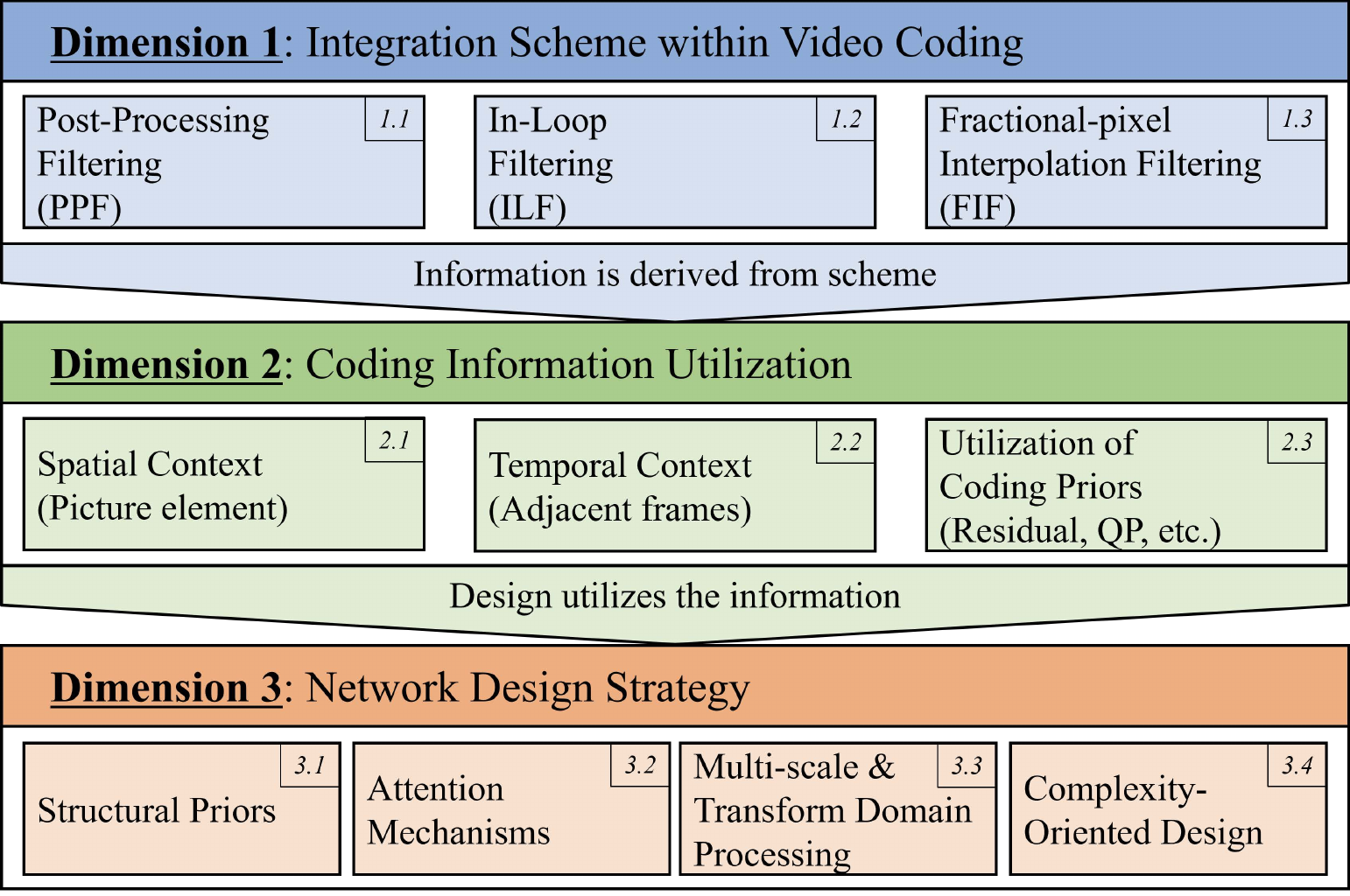}
	\caption{\scriptsize Proposed Systematic Three-dimensional Taxonomy of Deep Learning-based Filtering for Video Coding.}
	\label{fig:taxo}
\end{figure}

\IEEEpubidadjcol

		The success of Deep Learning (DL), especially Convolutional Neural Networks (CNNs) \cite{LeCun1989BackpropagationAT, LeCun1998GradientbasedLA}, has inspired substantial research on applying DL to video coding. 
		Dong \textit{et al.} early demonstrated that CNNs are not only suitable for Super-Resolution (SR), but also can effectively remove artifacts and reconstruct visual quality \cite{Dong2015SRCNN,Dong2015ARCNN}. 
		Following these advances, numerous studies have proposed DLF for video coding tasks, including Post-Processing Filtering (PPF), In-Loop Filtering (ILF), and Fractional-pixel Interpolation Filtering (FIF). 
		
		It is worth noting that the vast majority of these solutions were heavily concentrated on HEVC/H.265 and VVC/H.266.
		The main reason is that the explosive maturation of CNNs and general-purpose computing coincided exactly with the active deployment era of HEVC/H.265, making it the natural testbed for early DLFs.
		Subsequently, this momentum seamlessly transitioned into the JVET's VVC/H.266 standardization process.
		More recently, the JVET's NNVC \cite{NNVCAHG11, NNVCEE1} has further accelerated the integration of DL tools into coding pipeline.
		Most importantly, since HEVC/H.265 and VVC/H.266 represent the most dominant and actively deployed standards in the CE ecosystem from mobile to 8K display, research efforts naturally gravitate toward these frameworks to ensure viability.
		
		Despite these developments, existing literature primarily focuses on individual components, highlighting the need for a comprehensive survey that offers a unified, system-level analysis across the entire pipeline. 
		Furthermore, to bridge the gap between theory and practical deployment in CE, such a survey is essential to investigate stringent hardware constraints, including computational efficiency, i.e., Multiply-Accumulate operations (MACs), memory bandwidth limits (e.g., avoiding line-buffers), integer-only inference, i.e., Small Ad-hoc Deep Learning (SADL), and standardization challenges.
		
		This paper provides a taxonomy-driven overview of DLF specifically for video coding, offering both theoretical insights and practical perspectives relevant to HEVC/H.265, VVC/H.266, and emerging NNVC systems. 
		The main contributions of this work are fivefold and summarized below:

		\begin{itemize}
			\item\textbf{Systematic Three-dimensional Taxonomy for DLF:}
				As illustrated in Fig.~\ref{fig:taxo}, we organize the DLF design space along three axes that each govern a distinct CE deployment constraint: \textit{where} the filter is integrated (Dimension~1: Integration Scheme within Video Coding) determines pipeline latency and bitstream-syntax dependency; \textit{what} information the filter consumes (Dimension~2: Coding Information Utilization) determines parsing dependency and memory bandwidth; and \textit{how} the network is structured (Dimension~3: Network Design Strategy) determines per-pixel MACs and on-chip storage.
				The axes are independently variable: any placement (PPF/ILF/FIF), any input set (spatial, temporal, or coding prior), and any architecture (residual, attention, multi-scale, etc.) can be combined, and all such combinations appear in the surveyed literature.
				Compared to prior reviews, this taxonomy offers a more complete and video coding-aware framework for understanding the evolution of DLFs.
	
			\item\textbf{End-to-End (E2E) System Perspective Across the Coding Pipeline:}
				Whereas existing surveys focus on individual filtering stages, this work holistically analyzes DLF across PPF, ILF, and FIF, covering the representative methods from \cite{DaiY2017CNNPostHEVCIntra,LiC2017CNNPostHEVC,WangT2017DecoderEndHEVC,YangR2017ScalableCNNQE,LiF2018DRNQE,HeX2018PartitionMaskQE,MaL2018ResidualBasedQE,WangT2018MultiScaleDeepDecoder,YangR2018MultiFrameQE,SohJW2018DeepTemporalAR,FengL2019CodingPriorQE,Hoang2019BDRRN,MengX2019MGANet,MengX2019SDTSQE,TongJ2019MultiFrameQE,YangR2019EnhancingHEVC,YangR2019QualityGatedLSTM,YuL2019MRRNQE,LuM2019LearnedRestorationVVCIntra,ChenW2020NN_AR_Temporal,LiH2020QEVC,LiX2020MSGDN,LuG2020DeepNonLocalKalman,MengX2020BSTN,SunW2020QE_NoiseDist,WangJ2020MWGAN,WangT2020VisualPerceptionGAN,XingQ2020EarlyExitQE,ZhangF2020EnhancingVVCPost,ZhaoH2020CNNPostEfficiency,HuangH2021FrameWiseQE,Santamaria2021ContentAdaptivePost,DingD2022BipredQE,QiZ2022CNNPostMultiScaleDWT,SunW2022QE_CBR,Lan2023CNN,Das2023High,ZhangH2023WCDANN,ParkWS2016CNNILF,KangJ2017MultiModalILF,JiaC2017STResNetILF,WangY2018DenseResidualILF,ZhangY2018RHCNNILF,DaiY2018CNNILFCUClassHEVC,MengX2018LSTMResNetILF,SongX2018PracticalCNNILF,ChenG2019AV1ILF_WARN,DingD2019CNNILFAV1,JiaC2019ContentAwareILF,LiT2019MultiFrameILF,WangD2019PartitionTreeILF,WangM2019AttentionDualScaleILF,WangM2019IntegratedPostVVCIntra,XuX2019DIANetILF,DingD2020SwitchableILF,HuangZ2020MultiGradientILF,Lam2020EfficientAdaptation,PanZ2020EfficientILF_EDCNN,ZhangS2020RRCNNILF,HuangZ2021EfficientQPVarILF,Nasiri2021ModelSelectionQE,Nasiri2021PredictionAwareQE,WangZ2021MDCNNILF,Bouaafia2021DeepQE_VVC,MaD2021MFRNet,Li2022Neural,HuangZ2022OneForAll,Kathariya2022MultiStageLocally,Kathariya2022MultiStageSpatial,LimWQ2022ALF_CNNClass,LimWQ2022PerfCompALF_CNNClass,LiY2023iDAM,Zhao2022Joint,ZhangH2023Lightweight,Man2023Meta,Kathariya2024Joint,Cui2024PFTILF,TongO2024Swin,FengZ2024Low,ZhangH2024RTNN,Kim2024ReferencebasedIF,Zhu2024NeuralNB,Zhang2024ARF,Man2025ContentAwareDI,Zhao2025AdvancedLC,ZhangH2017LearningCNNFI,Xia2018SwitchModeFI,Ibrahim2018NNFME,Pham2019DeepFI,YanN2019CNNBasedFI,Murn2020InterpretingCF,Murn2021ImprovedCNNFI,Lee2024PixelEF,Li2025CCLOPCE,Li2025ScreenCV,Cho2025SpatialChannelMB,Qin2025CombinationTO,Zhu2024CPGA,Liu2024EMAFA,Yu2024MultiSwin,Das2024VVCPPFF,Wang2025STFF,HoangVan2025OVQEVVC,Gai2025DRGNet,Zeng2025PnPVCVE}.
				This E2E perspective highlights how DLF tools interact with prediction, reconstruction, and Motion Estimation/Compensation (ME/MC).
			
			\item\textbf{Comprehensive Complexity vs. Performance Analysis:}
				Using quantitative results from the literature including Bjøntegaard Delta rate (BD-rate) gains, computational complexity (kMACs), model size (number of parameters), and decoding time data, we present the first extended trade-off analysis (Fig.~\ref{fig:performance}) comparing HEVC Test Model (HM), VVC Test Model (VTM), and Enhanced Compression Model (ECM), across All-Intra (AI), Random Access (RA), and Low-Delay (LD) coding configurations.
				This offers practical guidance for real-time CE devices, where complexity is as critical as coding efficiency.	
			
			\item\textbf{Video Coding Standard-Oriented Discussion:}
				We incorporate the latest findings from JVET NNVC Ad-Hoc Group 11 (AHG11) activities and highlight obstacles such as floating-point reproducibility, fixed-point inference, parameter transmission, and model signaling. 
				This establishes the link between academic research and standardization.
			
			\item\textbf{Open Challenges and Future Directions:}
				Finally, we identify key unsolved issues including temporal flicker suppression \cite{SohJW2018DeepTemporalAR}, error propagation in ILF, domain generalization, dynamic complexity control \cite{XingQ2020EarlyExitQE, Man2025ContentAwareDI}, and NPU-friendly model design and present future research directions for next-generation intelligent video coding.
		\end{itemize}
		\noindent The remainder is organized as follows: Section \ref{background} covers the background; Section \ref{integration} reviews PPF, ILF, and FIF methods; Sections \ref{utilization}--\ref{structure} address coding-information utilization and network design; Section \ref{performance} analyzes complexity vs.\ performance; and Section \ref{challenges} discusses open challenges and future directions. Finally, Section \ref{conclusion} concludes this article.

\begin{table}[!t]
	\caption{\scriptsize Condensed Summary of Deep Learning-based Filtering}
	\label{tab:summary}
	\vspace{-8pt}
	\renewcommand{\arraystretch}{0.56}
	\scriptsize
	\centering
	\begin{tabular}{c|c|c|l}\hline\hline
		Ref. & Year & Std. / SW & \multicolumn{1}{c}{Primary Technique}\\\hline\hline
		\cite{DaiY2017CNNPostHEVCIntra} & 2017 &  HM-16.0 & Variable filter size, Residual learning \\
		\cite{LiC2017CNNPostHEVC} & 2017 &  x265 & Dynamic Metadata \\
		\cite{WangT2017DecoderEndHEVC} & 2017 &  HM-16.0 & Deep CNN (20 layers) \\
		\cite{YangR2017ScalableCNNQE} & 2017 &  HM-16.0 & Scalable Architecture \\
		\cite{LiF2018DRNQE} & 2018 &  HEVC & Global Residual Learning \\
		\cite{HeX2018PartitionMaskQE} & 2018 &  HM-16.0 & Input: CU Partition Mask \\
		\cite{MaL2018ResidualBasedQE} & 2018 &  HEVC & Input: Residual Signal \\
		\cite{WangT2018MultiScaleDeepDecoder} & 2018 &  HM-16.0 & ConvLSTM, Multi-scale \\
		\cite{YangR2018MultiFrameQE} & 2018 &  HEVC & Peak Quality Frames (PQFs) \\
		\cite{SohJW2018DeepTemporalAR} & 2018 &  HM-16.9 & Temporal Consistency Modeling \\
		\cite{FengL2019CodingPriorQE} & 2019 &  x265 & Input: Unfiltered \& Prediction \\
		\cite{Hoang2019BDRRN} & 2019 &  HM-20.0 & Recursive Residual, Block Info \\
		\cite{MengX2019MGANet} & 2019 &  HM-16.9 & Bidirectional LSTM, Guide Map \\
		\cite{MengX2019SDTSQE} & 2019 &  VTM-3.0 & Temporal Structure Fusion \\
		\cite{TongJ2019MultiFrameQE} & 2019 &  HM-16.9 & Optical Flow + Early Fusion \\
		\cite{YangR2019EnhancingHEVC} & 2019 &  HM-16.0 & Time-constrained Optimization \\
		\cite{YangR2019QualityGatedLSTM} & 2019 &  HM-16.0 & Quality-Gated LSTM \\
		\cite{YuL2019MRRNQE} & 2019 &  HM-16.12 & Recursive Residual \\
		\cite{LuM2019LearnedRestorationVVCIntra} & 2019 & VVC & Multi-scale Spatial Priors \\
		\cite{ChenW2020NN_AR_Temporal} & 2020 &  HM-16.0 & Deformable Convolution \\
		\cite{LiH2020QEVC} & 2020 &  HM-16.19 & Deep Reinforcement Learning \\
		\cite{LiX2020MSGDN} & 2020 &  VTM-8.0 & Multi-scale Grouped Dense, GAN \\
		\cite{LuG2020DeepNonLocalKalman} & 2020 &  x265 & Deep Kalman Filtering \\
		\cite{MengX2020BSTN} & 2020 &  HM-16.9 & FlowNet, Bi-LSTM, TU Map \\
		\cite{SunW2020QE_NoiseDist} & 2020 &  HM-16.0 & Noise Distribution Prior \\
		\cite{WangJ2020MWGAN} & 2020 &  HM-16.5 & Wavelet-based GAN \\
		\cite{WangT2020VisualPerceptionGAN} & 2020 &  HEVC & Visual Perception GAN \\
		\cite{XingQ2020EarlyExitQE} & 2020 &  HM-16.5 & Early-Exit Mechanism \\
		\cite{ZhangF2020EnhancingVVCPost} & 2020 &  VTM-4.0.1 & Residual Learning \\
		\cite{ZhaoH2020CNNPostEfficiency} & 2020 &  HM-16.19 & Batch Norm, Variable Filter \\
		\cite{HuangH2021FrameWiseQE} & 2021 &  HM-16.18 & Input: Mode Map \\
		\cite{Santamaria2021ContentAdaptivePost} & 2021 &  NNVC-1.0 & Bias-only Fine-tuning \\
		\cite{DingD2022BipredQE} & 2022 &  HM-16.9 & Bi-prediction (Virtual Frame) \\
		\cite{QiZ2022CNNPostMultiScaleDWT} & 2022 &  NNVC-1.0 & Discrete Wavelet Transform \\
		\cite{SunW2022QE_CBR} & 2022 &  HM-16.9,x265 & Coding Priors (CBR) \\
		\cite{Lan2023CNN} & 2023 & NNVC-2.0 & Single PPF for various distortion \\
		\cite{Das2023High} & 2023 & VVenC/VVdeC & QP-Map, Lightweight Network \\
		\cite{ZhangH2023WCDANN} & 2023 & NNVC & Depthwise Separable Conv. + Attn. \\
		\cite{Zhu2024CPGA} & 2024 & HM-16.25 & Temporal/Non-local Aggregation \\
		\cite{Liu2024EMAFA} & 2024 & HM-16.5 & Deformable Conv., Frequency-Aware \\
		\cite{Yu2024MultiSwin} & 2024 & HEVC/VTM-22.0 & Swin Transformer, Spatio-Temporal \\
		\cite{Das2024VVCPPFF} & 2024 & VVenC v1.10.0 & Feature Fusion CNN, QP-Adaptive \\
		\cite{Wang2025STFF} & 2025 & HM-16.5 & Spatio-Temporal + High-Freq. \\
		\cite{HoangVan2025OVQEVVC} & 2025 & VVenC & Omni-Freq. Adaptive \\
		\cite{Gai2025DRGNet} & 2025 & VTM & Dual-Path Gating, QP-map \\
		\cite{Zeng2025PnPVCVE} & 2025 & AVC/HEVC/VVC & Plug-and-Play, Bitstream-Aware \\\hline
		\cite{ParkWS2016CNNILF} & 2016 &  HM-16.0 & Shallow CNN, QP-specific \\
		\cite{KangJ2017MultiModalILF} & 2017 &  HM-16.7 & Multi-modal/Multi-scale \\
		\cite{JiaC2017STResNetILF} & 2017 &  HM-16.15 & Temporal Input \\
		\cite{WangY2018DenseResidualILF} & 2018 &  HM-16.15 & Dense Connection \\
		\cite{ZhangY2018RHCNNILF} & 2018 &  HM-12.0 & Residual Highway Unit \\
		\cite{DaiY2018CNNILFCUClassHEVC} & 2018 &  HM-16.0 & CU Classification \\
		\cite{MengX2018LSTMResNetILF} & 2018 &  HM-7.0 & LSTM + Residual \\
		\cite{SongX2018PracticalCNNILF} & 2018 &  JEM-7.0 & Input: QP Map, Fixed-point \\
		\cite{ChenG2019AV1ILF_WARN} & 2019 &  AV1 & Wide-Activation ResNet \\
		\cite{DingD2019CNNILFAV1} & 2019 &  AV1 & Depth-variable Network \\
		\cite{JiaC2019ContentAwareILF} & 2019 &  HM-16.9 & Content-Aware Models \\
		\cite{LiT2019MultiFrameILF} & 2019 &  HM-16.5 & Reference Frame Selector \\
		\cite{WangD2019PartitionTreeILF} & 2019 &  HM-16.15 & Input: Partition Tree Map \\
		\cite{WangM2019AttentionDualScaleILF} & 2019 &  VTM-4.0 & Dual-scale, Attention \\
		\cite{WangM2019IntegratedPostVVCIntra} & 2019 &  VTM-3.0 & Input: QP, Partition Mode \\
		\cite{XuX2019DIANetILF} & 2019 &  HM-16.20 & Dense Inception, Attention \\
		\cite{DingD2020SwitchableILF} & 2020 &  HM-16.9 & Switchable Mechanism \\
		\cite{HuangZ2020MultiGradientILF} & 2020 &  VTM-3.0 & Input: Image Gradients \\
		\cite{Lam2020EfficientAdaptation} & 2020 &  VTM-7.0 & Bias-only Fine-tuning \\
		\cite{PanZ2020EfficientILF_EDCNN} & 2020 &  HM-16.9 & Weighted Normalization \\
		\cite{ZhangS2020RRCNNILF} & 2020 &  HM-16.16 & Recursive Residual \\
		\cite{HuangZ2021EfficientQPVarILF} & 2021 &  VTM-6.0 & Input: QP Attention (QPAM) \\
		\cite{Nasiri2021ModelSelectionQE} & 2021 &  VTM-10.0 & Model Selection Signaling \\
		\cite{Nasiri2021PredictionAwareQE} & 2021 &  VTM-6.0 & Input: Prediction Signal \\
		\cite{WangZ2021MDCNNILF} & 2021 &  VTM-9.3 & Multi-density Block \\
		\cite{Bouaafia2021DeepQE_VVC} & 2021 & VTM-4.0 & U-Net-based Wide-activated SENet \\
		\cite{MaD2021MFRNet} & 2021 & VTM-7.0 & Multi-level Feature Review \\
		\cite{Li2022Neural} & 2022 & VTM-11.0 & Analysis of DLFs\\
		\cite{HuangZ2022OneForAll} & 2022 &  VTM-6.0 & Input: QP/FT Attention \\
		\cite{Kathariya2022MultiStageLocally} & 2022 &  VTM-11.0 & Dual-domain Fusion \\
		\cite{Kathariya2022MultiStageSpatial} & 2022 &  VTM-11.0 & Transformer \\
		\cite{LimWQ2022ALF_CNNClass} & 2022 &  VTM-13.0 & CNN-based Classification \\
		\cite{LimWQ2022PerfCompALF_CNNClass} & 2022 &  VTM-13.0 & Complexity Control \\
		\cite{LiY2023iDAM} & 2023 &  VTM-11.0 & Iterative Training \\
		\cite{Zhao2022Joint} & 2023 &  VTM-14.0 & Joint Luma-Chroma \\
		\cite{ZhangH2023Lightweight} & 2023 & NNVC-2.0 & Lightweighting similar to \cite{ZhangH2023WCDANN} \\
		\cite{Man2023Meta} & 2023 & VTM-11.0 & Dynamic weight, Meta-learning \\
		\cite{Kathariya2024Joint} & 2024 & VTM-11.0 & Pixel and DCT fusion \\
		\cite{Cui2024PFTILF} & 2024 & NNVC-4.0 & Partition-level Transformer \\
		\cite{TongO2024Swin} & 2024 & VTM-11.0 & Swin Transformer \\
		\cite{FengZ2024Low} & 2024 & NNVC-2.0 & Feature Distillation, Asymmetric Conv. \\
		\cite{ZhangH2024RTNN} & 2024 & NNVC-3.0 & ResNet + Transformer \\
		\cite{Kim2024ReferencebasedIF} & 2024 &  VVC & Ref-to-Current Estimation \\
		\cite{Zhu2024NeuralNB} & 2024 &  VTM-11.0 & Multi-level Framework \\
		\cite{Zhang2024ARF} & 2024 &  VTM-11.0 & Reconfigurable Framework \\
		\cite{Man2025ContentAwareDI} & 2025 &  VTM-11.0 & Dynamic Network Topology \\
		\cite{Zhao2025AdvancedLC} & 2025 &  ECM-14.0 & NN-based Intra Prediction \\
		\cite{Li2025CCLOPCE} & 2025 & NNVC-10 & Cross-Component LOP Filter \\
		\cite{Li2025ScreenCV} & 2025 & HM-16.20 & Auxiliary Maps, Non-local Models\\
		\cite{Cho2025SpatialChannelMB} & 2025 & NNVC-13.0 & Spatial-Channel Mixing (SCM) Block \\
		\cite{Qin2025CombinationTO} & 2025 & NNVC-10.0 & Test of NNVC tools, TRFS-based \\\hline
		\cite{ZhangH2017LearningCNNFI} & 2017 &  HM-16.7 & SRCNN-like Architecture \\
		\cite{Xia2018SwitchModeFI} & 2018 &  HM-16.7 & H-/Q-pel Branch, CU-level Switching\\
		\cite{Ibrahim2018NNFME} & 2018 &  HM-16.9 & 4-layer CNN \\	
		\cite{Pham2019DeepFI} & 2019 &  HM-16.18 & Luma \& Chroma \\
		\cite{YanN2019CNNBasedFI} & 2019 &  HM-16.7 & Regression Problem \\
		\cite{Murn2020InterpretingCF} & 2020 &  HM-16.20 & Explainability \\
		\cite{Murn2021ImprovedCNNFI} & 2021 &  VTM-6.0 & Residual + Attention \\
		\cite{Lee2024PixelEF} & 2024 &  VTM-11.0 & Pixel-wise Estimation \\
		\hline\hline
	\end{tabular}
	\vspace{-16pt}
\end{table}

	\section{Background}\label{background}
		\subsection{Hybrid Video Coding Framework and Visual Artifacts}
			Modern standards like VVC/H.266 adhere to a block-based hybrid coding architecture.  
			Input frames are partitioned into Coding Tree Units (CTUs) and processed via prediction, transform, and quantization. 
			While efficient, the independent quantization of transform coefficients inevitably introduces artifacts: \textit{blocking} (discontinuities at boundaries), \textit{ringing} (oscillations near edges), and \textit{blurring} (loss of high-frequency detail).

		\subsection{Conventional Video Coding Filter}
			To alleviate the artifacts generated during the coding process, video coding standards incorporate a suite of filters.
			\begin{itemize}
				\item \textit{Deblocking Filter (DBF)}: 
				Operates on block boundaries to reduce blocking artifacts using a low-pass filter based on boundary strength and Quantization Parameters (QPs).
				
				\item \textit{Sample Adaptive Offset (SAO)}:
				Compensates for pixel intensity offsets by classifying pixels into edge or band types, reducing ringing and banding artifacts.
				
				\item \textit{Adaptive Loop Filter (ALF)}:
				Minimizes the Mean Squared Error (MSE) between the original and reconstructed signals using Wiener-based linear filter with adaptive coefficients signaled in the bitstream.
				
				\item \textit{Luma Mapping with Chroma Scaling (LMCS)}:
				Remaps luma signals to redistribute coding noise and scales chroma residuals based on luma dependencies to improve efficiency.
				
				\item \textit{Discrete Cosine Transform-based Interpolation Filter (DCTIF)}:
				Uses Finite Impulse Response (FIR) filters derived from Discrete Cosine Transform (DCT) to generate fractional-pixel samples for ME/MC.
			\end{itemize}
			\noindent Despite their effectiveness, these filters are limited by their linear nature and small parameter sets, struggling to model complex non-linear signal dependencies.

		\subsection{Neural Network-based Video Coding}
			To standardize data-driven approaches, the JVET NNVC explores replacing core modules with Neural Network (NN) techniques \cite{NNVCAHG11,NNVCEE1}. 
			Unlike E2E learning schemes, the standardization effort primarily focuses on a hybrid approach where lightweight NN models augment specific block-based coding tools.
			Key areas include NN-based Intra Prediction (NNIP) for complex texture modeling \cite{kim2022nn}, NN-based In-Loop Filter (NNILF) for artifact removal \cite{wang2022ee1}, and SR-based Reference Picture Resampling (RPR) for spatial scalability \cite{schorn2021dictionary,fu2023cnn}.
			Specifically, these candidates are evaluated under strict complexity constraints, such as the maximum number of multiplications per pixel and parameter count limits, to ensure feasibility.
			A critical development is the SADL framework, which supports integer-based inference.
			By enforcing fixed-point arithmetic, SADL guarantees bit-exact cross-platform reproducibility, a mandatory requirement for international standards, thereby directly addressing the computational complexity and power constraints of CE silicon, including mobile System-on-Chip (SoC) platforms, NPUs, and edge devices, where integer dataflow is the dominant execution mode.

	\section{Integration Scheme within Video Coding}\label{integration}
		Following Dimension 1 of our proposed taxonomy as illustrated in Fig. \ref{fig:taxo}, this section analyzes DLF methods by their integration location within the hybrid video coding framework: PPF, ILF, and FIF.
		Each placement choice imposes a distinct hardware-cost profile (latency, syntax dependency, error-propagation risk), which is the primary reason placement is treated as a separate dimension rather than a sub-attribute of network design.
		TABLE~\ref{tab:summary} summarizes the surveyed methods; we focus on structural advantages and practical trade-offs of each approach.
		It is divided into PPF (top), ILF (middle), and FIF (bottom).
		
		\textbf{Scope and Comparability Statement:} It is important to note that these three schemes operate under fundamentally different constraints, and their quantitative gains and complexity overheads are not directly comparable. 
		\textbf{PPF} operates only at the decoder side as a post-processor, focusing on visual enhancement without affecting the coding loop. Its complexity is purely an add-on to the decoder.
		\textbf{ILF} is embedded within the coding loop, where filtered frames serve as reference frames. While it offers coding gain by improving prediction accuracy, it imposes complexity on both the encoder and decoder and carries the risk of error propagation.
		\textbf{FIF} replaces the interpolation filter in the ME/MC module. It is invoked frequently at the block level, making its per-pixel complexity constraints significantly stricter than frame-level filters.
		Therefore, the performance metrics presented in subsequent sections should be interpreted within the specific context of each integration.

		\subsection{Post-Processing Filtering (PPF)}
			\subsubsection{Single-Frame Enhancement}
				Early research primarily focused on recovering high-frequency components lost due to quantization using CNNs.
				Dai \textit{et al.} \cite{DaiY2017CNNPostHEVCIntra} and Li \textit{et al.} \cite{LiF2018DRNQE} effectively suppressed blocking and ringing artifacts in HEVC/H.265 intra frames through variable filter sizes and residual learning.
				However, these methods treat the entire frame uniformly, failing to differentiate between artifact-heavy block boundaries and smooth regions, which leads to redundant computations.	
				To overcome this, methods utilizing coding information have emerged.
				He \textit{et al.} \cite{HeX2018PartitionMaskQE} and Huang \textit{et al.} \cite{HuangH2021FrameWiseQE} leveraged partition information such as Coding Unit (CU) and Transform Unit (TU) or prediction mode maps to guide the network toward artifact-prone regions, thereby enhancing restoration efficiency.
				However, these approaches require extracting internal coding information, increasing implementation dependency compared to pure `blind post-processing'.
				Furthermore, generative methods such as Multi-Scale Grouped Dense Network \cite{LiX2020MSGDN} excel in perceptual quality but are controversial from a coding efficiency perspective due to lower pixel-wise accuracy and the risk of generating hallucination artifacts.
		
			\subsubsection{Multi-Frame Enhancement}
				Multi-frame approaches that exploit temporal correlations in video generally outperform single-frame methods.
				Yang \textit{et al.} mitigated quality fluctuation issues by detecting Peak Quality Frames \cite{YangR2018MultiFrameQE}.
				However, the process of identifying and aligning them introduces high latency.
				Lu \textit{et al.} \cite{LuG2020DeepNonLocalKalman} and Yang \textit{et al.} \cite{YangR2019QualityGatedLSTM} adopted time series structures such as Recurrent Neural Networks (RNNs) and Long Short-Term Memory (LSTM) networks to efficiently propagate temporal information, but this hinders parallel processing and significantly increases memory bandwidth requirements.
				Recently, Ding \textit{et al.} proposed Progressive Motion-compensated Video Enhancement \cite{DingD2022BipredQE} for synthesizing virtual frames via bi-prediction instead of complex optical flow estimation to reduce computational load, yet it still requires buffering multiple frames, making it unsuitable for low-delay applications.
				Soh \textit{et al.} also addressed temporal inconsistency by proposing a deep temporal network to reduce flickering artifacts \cite{SohJW2018DeepTemporalAR}.
				Others like Tong \textit{et al.} utilized learning-based optical flow followed by early fusion to extract spatio-temporal features \cite{TongJ2019MultiFrameQE}, and Meng \textit{et al.} proposed a network to jointly exploit Spatial Details and Temporal Structure \cite{MengX2019SDTSQE}.

\begin{figure}[!t]
	\centering
	\includegraphics[width=\columnwidth]{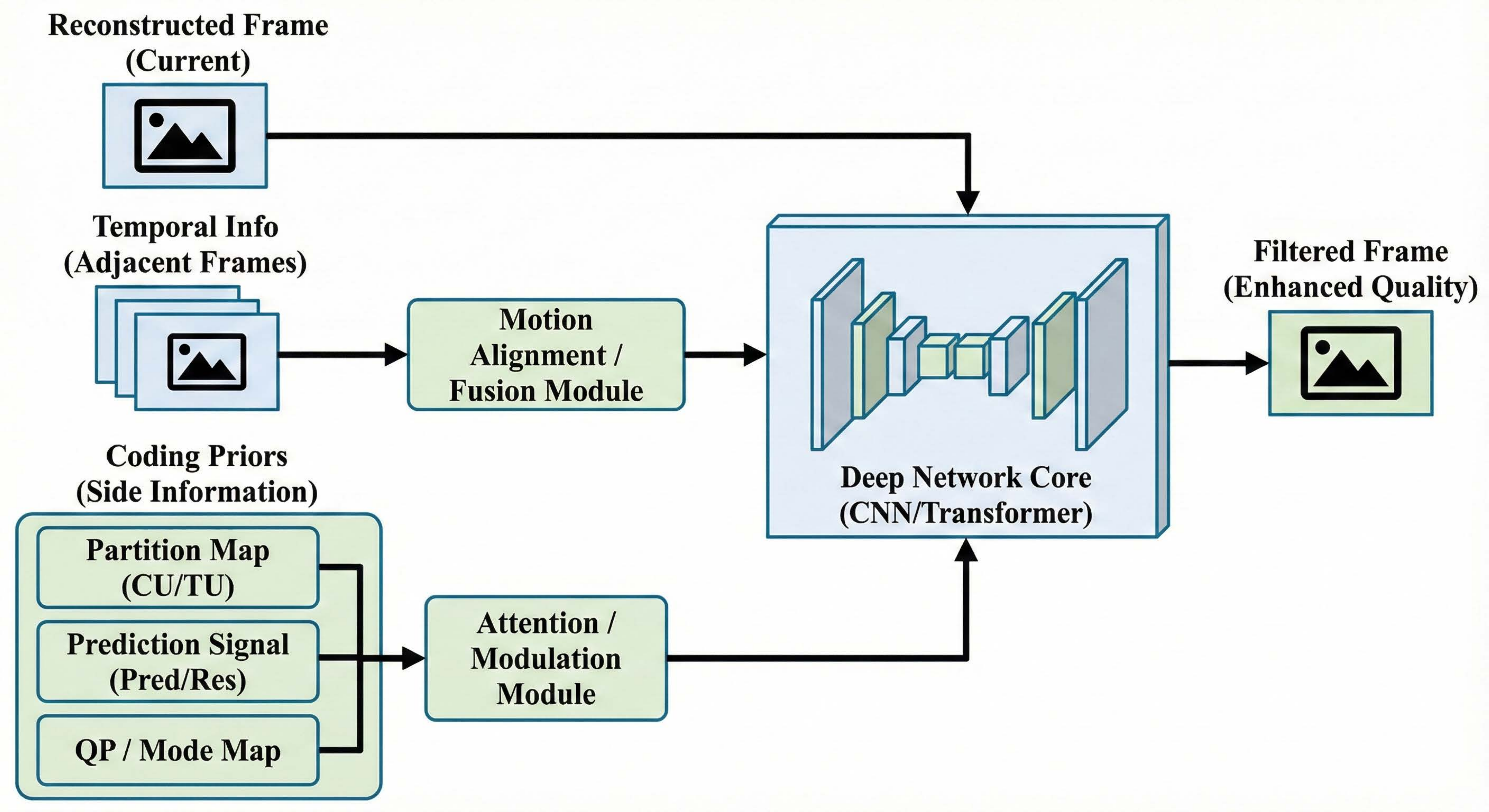}
	\caption{\scriptsize Conceptual Diagram of Utilizing Coding Information: Video coding utilizes spatial/temporal context and information derived from the  compression process.}
	\label{fig:utilization}
\end{figure}
		
			\subsubsection{Recent Advances}
				Recent works extended PPF to VVC/H.266 and video coding-agnostic settings.
				For HEVC/H.265, Zhu \textit{et al.} leveraged coding priors to guide multi-frame aggregation frameworks \cite{Zhu2024CPGA}.
				To improve high-frequency details, Liu \textit{et al.} incorporated deformable convolutions alongside frequency-aware components to adaptively mitigate structured artifacts \cite{Liu2024EMAFA}.
				Furthermore, Yu \textit{et al.} \cite{Yu2024MultiSwin} and Wang \textit{et al.} \cite{Wang2025STFF} advanced spatio-temporal fusion mechanisms, effectively blending spatial textures with temporal consistency maps.	
				For VVC/H.266,
				Das \textit{et al.} developed a QP-adaptive CNN framework without requiring multiple models for different compression rates \cite{Das2024VVCPPFF}.
				Hoang \textit{et al.} suggested a deep quality modeling \cite{HoangVan2025OVQEVVC}, while Gai \textit{et al.} introduced a Dual-path Residual Gating Network through streamlined feature propagation \cite{Gai2025DRGNet}.
				To maximize cross-standard utility, Zeng \textit{et al.} proposed a plug-and-play video quality enhancement network \cite{Zeng2025PnPVCVE} that unifies post-processing across H.264/AVC, HEVC/H.265, and VVC/H.266 by passively parsing bitstream metadata, thereby providing a highly flexible solution for heterogeneous CE devices.
	
		\subsection{In-Loop Filtering (ILF)}
			\subsubsection{Replacement vs. Enhancement Schemes}
				Approaches that replace traditional filters (DBF, SAO, ALF) with DL models, such as Park \textit{et al.} \cite{ParkWS2016CNNILF} and Zhang \textit{et al.} \cite{ZhangS2020RRCNNILF}, simplify the pipeline but lack a fail-safe mechanism if the DL model fails.
				Attempts like Multi-Gradient Non-Local Filter \cite{HuangZ2020MultiGradientILF} utilized image gradient information to preserve structural details, yet computational complexity remains high.
				Conversely, enhancement schemes that add DL modules after existing filters, such as Zhang \textit{et al.} \cite{ZhangY2018RHCNNILF} and Wang \textit{et al.} \cite{WangY2018DenseResidualILF}, ensure performance but significantly increase decoder complexity.
				Lim \textit{et al.} proposed generalizing the ALF with a CNN-based classifier to weight FIR filter outputs, enhancing performance \cite{LimWQ2022ALF_CNNClass}.
				Recent studies attempt hybrid and adaptive approaches to resolve this dilemma.
				Ma \textit{et al.} maximized performance through multi-level feature review but at the cost of high complexity \cite{MaD2021MFRNet}.
				In contrast, Li \textit{et al.} addressed content propagation issues by introducing iterative training and allowing block-level filter selection to prevent over-smoothing \cite{LiY2023iDAM}.
				Zhu \textit{et al.} proposed a multi-level ILF framework (Reference pixel, CTU, Frame levels) to further boost performance \cite{Zhu2024NeuralNB}.

\begin{table}[t]
	\centering
	\caption{\tiny{Signal Traceability Matrix: Mapping Input to Schemes, Granularity, and Signaling Overhead}}
	\label{tab:traceability}
	\resizebox{\linewidth}{!}{
		\begin{tabular}{l|l|c|c|c|l|l|l}
			\hline\hline
			\multirow{2}{*}{Domain} & 
			\multirow{2}{*}{Input Signal/Prior} & 
			\multicolumn{3}{c|}{Scheme Usage} & 
			\multirow{2}{*}{Granularity} & 
			\multirow{2}{*}{Availability Mode} & 
			\multirow{2}{*}{Signaling Overhead} 
			\\ 
			\cline{3-5}
			&  
			& 
			ILF & 
			FIF & 
			PPF &  
			&  
			&  
			\\\hline\hline
			\multirow{2}{*}{Pixel} & 
			Reconstructed & 
			$\bullet$ & 
			$\bullet$ & 
			$\bullet$ & 
			Frame/Patch & 
			Implicit(Recon.) & 
			None(Derived) 
			\\ 
			\cline{2-8} 
			& 
			Prediction Signal & 
			$\bullet$ & 
			$\bullet$ & 
			$-$ & 
			Block(PU) & 
			Implicit(Inter/Intra) & 
			None(Derived) 
			\\\hline
			\multirow{3}{*}{Coding} & 
			QP & 
			$\bullet$ & 
			$\bullet$ & 
			$\bullet$ & 
			Frame/Slice/CU & 
			Implicit(Parsed) & 
			None(But Parsing Dep.) 
			\\ 
			\cline{2-8} 
			& 
			Partition & 
			$\bullet$ & 
			$-$ & 
			$\circ$ & 
			CTU/Block & 
			Implicit(Parsed) & 
			None(But Parsing Dep.) 
			\\ 
			\cline{2-8} 
			& 
			Prediction Mode(I/P/B) & 
			$\bullet$ & 
			$\bullet$ & 
			$-$ & 
			Block(PU) & 
			Implicit(Parsed) & 
			None(But Parsing Dep.)
			\\\hline
			\multirow{2}{*}{Temporal} & 
			Motion Vectors & 
			$\circ$ & 
			$\bullet$ & 
			$-$ & 
			Block(PU) & 
			Implicit(Parsed) & 
			None(Derived) 
			\\ 
			\cline{2-8} 
			& 
			Reference Frame & 
			$\circ$ & 
			$\bullet$ & 
			$\circ$ & 
			Frame(Buffer) & 
			Implicit(DPB) & 
			None(High Bandwidth) 
			\\\hline
			\multirow{2}{*}{Control} & 
			Model/Filter Index & 
			$\bullet$ & 
			$\bullet$ & 
			$-$ & 
			Slice/CTU & 
			Explicit(Signaled) & 
			Low(Header/SEI) 
			\\
			\cline{2-8} 
			& 
			On/Off Control Flags & 
			$\bullet$ & 
			$\bullet$ & 
			$-$ & 
			CTU/Block & 
			Explicit(Signaled) & 
			High(Per-block bits) 
			\\\hline\hline
		\end{tabular}
	}
	\\[5pt]
	\tiny{$\bullet$: Primary Usage, $\circ$: Secondary/Optional Usage, $-$: Rarely Used. \\
		\textbf{Implicit}: Derived from bitstream (no extra bits). \textbf{Explicit}: Requires dedicated syntax elements.}
\end{table}
		
			\subsubsection{Complexity Control and Adaptability}
				A major criticism of ILF is the need to store multiple models to handle various QPs and frame types (I/P/B).
				To address this, Huang \textit{et al.} proposed efficient solutions using QP and frame types as conditional inputs (e.g., via attention mechanisms) to cover all conditions with a single model \cite{HuangZ2022OneForAll,HuangZ2021EfficientQPVarILF}.
				Kim \textit{et al.} enhanced robustness to different QPs in reference-based filtering using QP-aware convolutions \cite{Kim2024ReferencebasedIF}.
				Furthermore, Man \textit{et al.} \cite{Man2025ContentAwareDI} and Zhang \textit{et al.} \cite{Zhang2024ARF} introduced dynamic neural network techniques that dynamically alter network structures or prune layers based on input characteristics.
				This overcomes the inefficiency of static CNNs requiring fixed computation and suggests a crucial direction for next-generation video coding by enabling flexible control tailored to hardware resources.
			
			\subsubsection{Recent Advances}
				Recently, a key research focus is on lightweight model design to reduce computational complexity for practical hardware implementation and real-time processing. 
				For instance, a lightweight ILF tailored for real-time video coding has been proposed \cite{Zhao2025LightweightLI}. 
				In the JVET NNVC, a Cross-Component enhanced Low-complexity Operating Point (CCLOP) filter was explored, which incorporates deep luma features to improve chroma filtering performance \cite{Li2025CCLOPCE}.
				Alongside complexity reduction, structural optimizations of the ILF are continuously advancing. 
				A notable example is the integration of the Spatial-Channel Mixing block, which explicitly separates and sequentially applies spatial and channel mixing operations in NNILF backbones \cite{Cho2025SpatialChannelMB}. 
				Furthermore, addressing the growing demand for remote desktops and online meetings, a screen content-aware ILF method has been developed. 
				This approach utilizes non-local models and various prior maps to effectively compensate for distortion patterns specific to Screen Content Coding (SCC) \cite{Li2025ScreenCV}.
				Moreover, ILFs are increasingly being investigated from the perspective of joint optimization and synergy with other NN-based coding tools. 
				Comprehensive combination tests in VVC/H.266 and the ECM demonstrate that integrating NNILF alongside NN-based prediction can effectively exploit their complementarity, maximizing the overall coding gains for future video coding standards \cite{Qin2025CombinationTO, Zhao2025AdvancedLC}.

\begin{figure}[!t]
	\centering
	\includegraphics[width=\columnwidth]{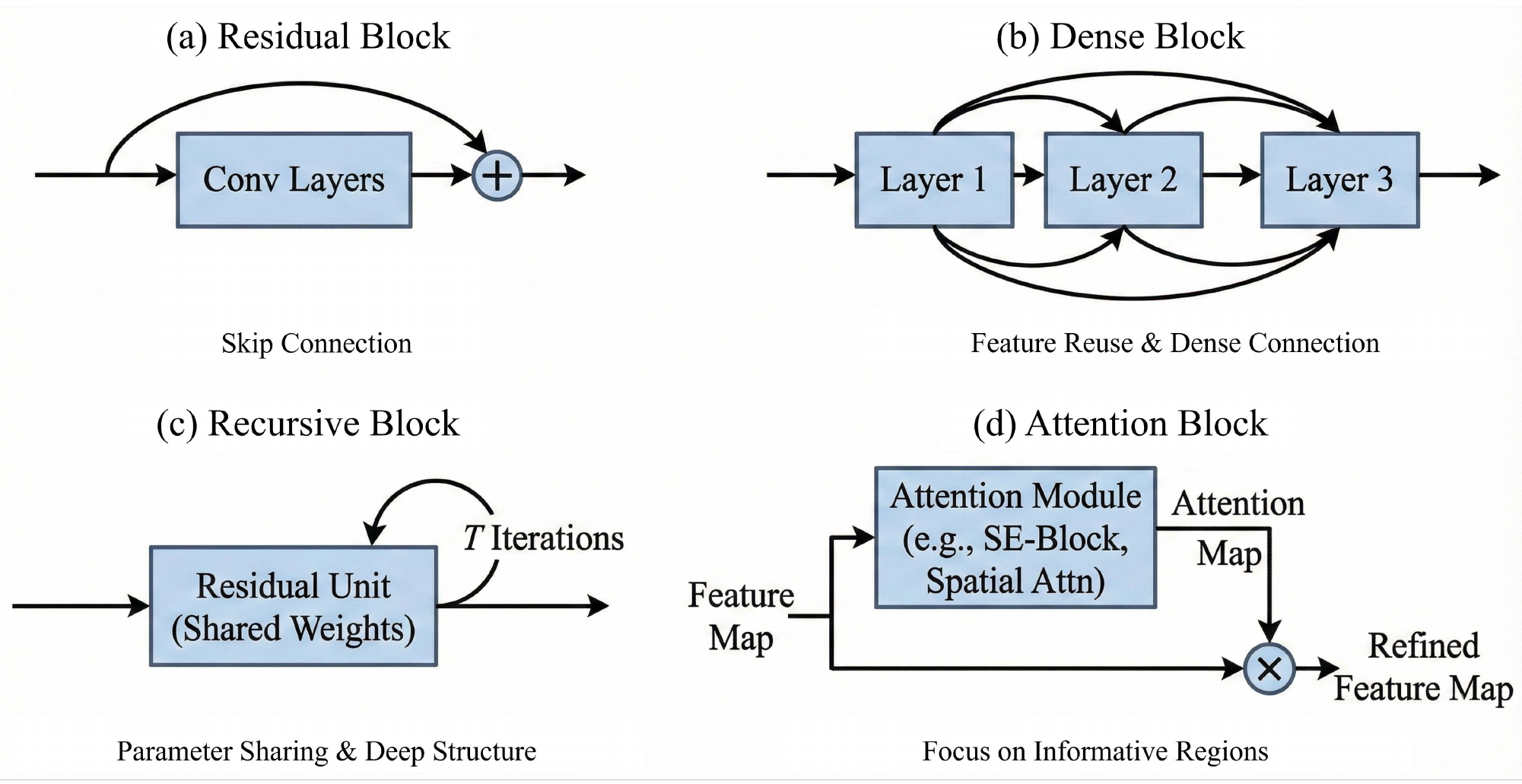}
	\caption{\scriptsize Key Building Blocks in Deep Neural Network Architectures for Video Filtering. (a) Residual Block, (b) Dense Block, (c) Recursive Block, and (d) Attention Block.}
	\label{fig:structure}
\end{figure}
			
		\subsection{Fractional-pixel Interpolation Filtering (FIF)}
			Studies replacing interpolation filters in the ME/MC stage have shown excellent efficacy in reducing prediction residuals.
			Yan \textit{et al.} reformulated interpolation as a regression problem, outperforming traditional DCTIF \cite{YanN2019CNNBasedFI}.
			Zhang \textit{et al.} also demonstrated the superiority of CNN-based interpolation over fixed filters \cite{ZhangH2017LearningCNNFI}.
			However, since ME/MC is the most frequently called module in video coding, applying CNNs introduces explosive computational overhead.
			To mitigate this, Murn \textit{et al.} analyzed the frequency response of CNN filters to design explainable lightweight models \cite{Murn2020InterpretingCF} and later improved the scheme for VVC/H.266 with attention mechanisms \cite{Murn2021ImprovedCNNFI}.
			Ibrahim \textit{et al.} proposed an interpolation-free scheme for HEVC/H.265 to reduce complexity \cite{Ibrahim2018NNFME}.
			Recently, Lee \textit{et al.} proposed a lightweight network that estimates filter coefficients pixel-wise to bypass heavy CNN computations \cite{Lee2024PixelEF}.
			The FIF domain has the highest increase in computation relative to quality gain; thus, practical implementation requires extreme lightweighting or integration with hardware accelerators.
			Pham and Zhou extended deep learning interpolation to both Luma and Chroma components, emphasizing completeness \cite{Pham2019DeepFI}.

	\section{Coding Information Utilization}\label{utilization}
		In accordance with Dimension 2 of Fig. \ref{fig:taxo}, this section explores how various information sources are utilized to enhance filtering performance.
		Fig. \ref{fig:utilization} conceptually summarizes the composition of training data and flow for the DLF.
		A key differentiator of DLF for video coding lies in the diversity of available information.
		Unlike image restoration tasks, video coding systems provide not only temporal correlations but also rich side information generated within the video coding process.
		
		Additionally, Table \ref{tab:traceability} provides a systematic traceability matrix mapping input priors to their target schemes, focusing on processing granularity and signaling overhead classification. 
		Crucially, the table distinguishes between `Implicit' priors (e.g., QP, Partition Info) that are derived from the bitstream without extra bits yet introduce parsing dependencies, and `Explicit' controls (e.g., On/Off flags) that incur direct bitrate overhead. 
		Furthermore, identifying signals requiring heavy `buffering' (e.g., Temporal Reference) versus those simply `parsed' is critical for estimating hardware constraints such as memory bandwidth and line-buffer requirements.
		
		\subsection{Spatial Context}
			Early research primarily focused on utilizing spatial information within a single frame.
			This approach restores textures lost due to quantization by exploiting correlations between adjacent pixels.
			Models such as Dai \textit{et al.} \cite{DaiY2017CNNPostHEVCIntra} and Li \textit{et al.} \cite{LiF2018DRNQE} used the reconstructed frame as input to remove blocking and ringing artifacts.
			To further enhance feature representation, Zhang \textit{et al.} utilized residual blocks to facilitate deeper network training without gradient vanishing \cite{ZhangY2018RHCNNILF}. 
			While this approach is simple to implement and effective for intra-coded frames, it is limited by overlooking the temporal characteristics of video and the structural information of the video coding.

		\subsection{Temporal Context}
			\subsubsection{Multi-Frame Fusion}
				To exploit the temporal redundancy inherent in video data, strategies fusing information from adjacent frames have been proposed.
				Yang \textit{et al.} exploited the quality fluctuation phenomenon as a reference to enhance lower-quality frames \cite{YangR2018MultiFrameQE}.
				Tong \textit{et al.} employed optical flow for motion compensation followed by an early fusion strategy to extract spatio-temporal features \cite{TongJ2019MultiFrameQE}.
		
			\subsubsection{Recurrent and Alignment-based Approaches}
				Yang \textit{et al.} selectively retained useful temporal information through quality gates \cite{YangR2019QualityGatedLSTM}, while Lu \textit{et al.} incorporated Kalman filtering theory to perform recursive restoration \cite{LuG2020DeepNonLocalKalman}.
				Additionally, Ding \textit{et al.} efficiently utilized temporal information by synthesizing virtual frames via bi-prediction instead of complex optical flow estimation \cite{DingD2022BipredQE}.
				Soh \textit{et al.} also proposed a deep temporal network to specifically address temporal flickering artifacts \cite{SohJW2018DeepTemporalAR}.
				Kim \textit{et al.} proposed a reference-based ILF that utilizes a reference-to-current feature estimation module to accurately predict textures from reference blocks \cite{Kim2024ReferencebasedIF}.

		\subsection{Utilization of Coding Priors}		
			\subsubsection{Partition Information}
				Since blocking artifacts occur at block boundaries, partition information serves as a crucial cue.
				He \textit{et al.} converted CU partition information into a mask to guide the network to focus on boundary regions \cite{HeX2018PartitionMaskQE}, and Wang \textit{et al.} utilized the multi-scale mean value of partitions as a guided map \cite{WangD2019PartitionTreeILF}.
				Meng \textit{et al.} further improved restoration performance by using boundary-guided maps derived from TU partition information \cite{MengX2019MGANet, MengX2020BSTN}.
			
			\subsubsection{Prediction Signals and Residuals}
				Ma \textit{et al.} proposed using residual signals alongside reconstructed frames as inputs, providing direct cues for texture restoration \cite{MaL2018ResidualBasedQE}.
				Feng \textit{et al.} defined unfiltered decoded frames and prediction frames as coding priors \cite{FengL2019CodingPriorQE}, and Nasiri \textit{et al.} similarly utilized prediction signals to enhance VVC/H.266 quality \cite{Nasiri2021PredictionAwareQE}.
			
			\subsubsection{QP and Mode Information}
				QP is a direct indicator of distortion levels.
				Song \textit{et al.} added a QP-map as an input channel to adapt a single model to various compression rates \cite{SongX2018PracticalCNNILF}.
				Advancing this, Huang \textit{et al.} proposed using QP and frame type information as inputs to an attention module to dynamically recalibrate channel-wise features \cite{HuangZ2021EfficientQPVarILF, HuangZ2022OneForAll}.
				Huang \textit{et al.} also employed a Mode Map combining intra-prediction modes and block sizes to effectively remove directional artifacts \cite{HuangH2021FrameWiseQE}.
				Furthermore, Man \textit{et al.} introduced a policy network that dynamically determines the filtering network structure based on pixel information and QP \cite{Man2025ContentAwareDI}.
				Sun \textit{et al.} incorporated noise distribution characteristics as a prior to guide the quality enhancement process \cite{SunW2020QE_NoiseDist}, and also proposed using coding priors for Constant Bit Rate (CBR) video coding scenarios \cite{SunW2022QE_CBR}.

\begin{figure}[!t] 
	\centering
	\includegraphics[width=\columnwidth]{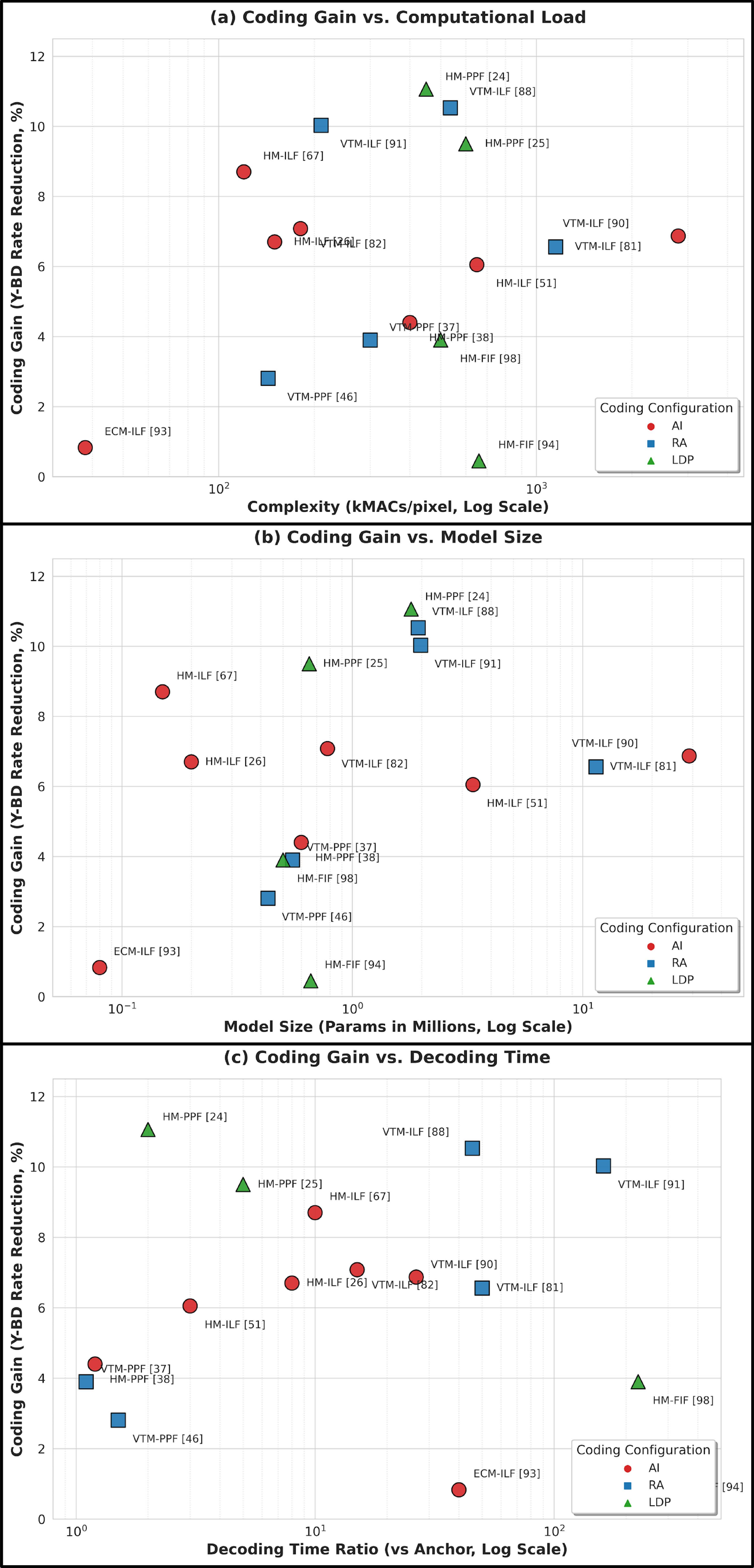}
	\caption{\scriptsize This scatter plot visualizes the evolution of DLFs, mapping coding gain (BD-rate) against computational complexity (kMACs/Params./Dec.time) on a log scale.}
	\label{fig:performance}
\end{figure}

	\section{Network Design Strategy}\label{structure}
		Based on Dimension 3 of Fig. \ref{fig:taxo}, we discuss the network design strategies and structural innovations employed in State-Of-The-Art (SOTA) DLF models.
		To effectively tackle the complex and non-linear nature of video compression artifacts, various architectural building blocks have been adopted, as illustrated in Fig. \ref{fig:structure}.
		The Residual Block (a) introduces skip connections to model the sparse difference (residual) between distorted and original signals, facilitating the training of deep networks by mitigating gradient vanishing.
		The Dense Block (b) maximizes feature reuse by connecting all layers, ensuring that hierarchical features extracted at different stages are preserved for robust restoration.
		To achieve a large receptive field with high parameter efficiency, the Recursive Block (c) iteratively applies shared weights, making it suitable for resource-constrained devices.
		Lastly, the Attention Block (d) explicitly models inter-dependencies between channels or spatial locations, allowing the network to dynamically concentrate its processing capacity on artifact-prone regions, such as complex textures and block boundaries, rather than treating all areas uniformly.

		\subsection{Structural Priors: Residual, Dense, and Recursive}
			\subsubsection{Global \& Local Residuals}
				To address the gradient vanishing problem in deep networks and to model the residual nature of compression artifacts (i.e., the difference between the original and distorted image is sparse), Residual Learning has become the standard.
				Dai \textit{et al.} \cite{DaiY2017CNNPostHEVCIntra} and Li \textit{et al.} \cite{LiF2018DRNQE} adopted global residual learning to force the network to predict the artifact map directly.
				Wang \textit{et al.} incorporated residual learning within dense blocks to facilitate feature reuse and gradient propagation \cite{WangY2018DenseResidualILF}.
			
			\subsubsection{Dense Connections}
				Dense connections allow feature reuse and improved gradient propagation by connecting all layers.
				Ma \textit{et al.} applied dense connections within its multi-level feature review blocks to propagate shallow features to deeper layers \cite{MaD2021MFRNet}.
				Similarly, Xu \textit{et al.} combined dense structures with inception blocks to preserve multi-scale information \cite{XuX2019DIANetILF}.
			
			\subsubsection{Recursive Structures}
				To increase the receptive field without significantly increasing the number of parameters, recursive networks share weights across layers.
				Zhang \textit{et al.} proposed a structure, which iteratively applies the same residual block to deepen the network while keeping the model size compact.
				Similarly, Hoang and Zhou utilized a recursive residual design in block-constrained structure to efficiently process block boundary information with shared weights \cite{Hoang2019BDRRN}.
				Yu \textit{et al.} extended this concept to handle artifacts at different scales iteratively \cite{YuL2019MRRNQE}.	

		\subsection{Attention Mechanisms}
			\subsubsection{Channel Attention}
				Ding \textit{et al.} integrated attention blocks to explicitly model the inter-dependencies between channels, emphasizing feature maps that contain relevant artifact information \cite{DingD2020SwitchableILF}.
				Huang \textit{et al.} proposed an attention module for QP to re-weight channel features dynamically, allowing a single model to adapt to different quality levels \cite{HuangZ2021EfficientQPVarILF, HuangZ2022OneForAll}.
			
			\subsubsection{Spatial \& Mixed Attention}
				Xu \textit{et al.} combined dense inception blocks with both spatial and channel attention to capture multi-scale features and focus on artifact-prone spatial locations \cite{XuX2019DIANetILF}.
				Wang \textit{et al.} also employed a dual-scale attention mechanism to refine features at different resolutions simultaneously \cite{WangM2019AttentionDualScaleILF}.
				Zhao \textit{et al.} utilized the Convolutional Block Attention Module (CBAM) to effectively learn correlations between luma and chroma components \cite{Zhao2022Joint}.
					
		\subsection{Multi-scale and Transform Domain Processing}
			\subsubsection{Multi-scale Architectures}
				Kang \textit{et al.} \cite{KangJ2017MultiModalILF} and Wang \textit{et al.} \cite{WangT2018MultiScaleDeepDecoder} utilized downsampling to extract features at different resolutions, effectively capturing global context and local details.
				Li \textit{et al.} designed a multi-scale grouped dense network to extract features at varying scales for better restoration \cite{LiX2020MSGDN}.
			
			\subsubsection{Transform Domain}
				Qi \textit{et al.} utilized Discrete Wavelet Transform (DWT) to decompose frames into sub-bands, processing them with a step-like network to separate high-frequency artifacts from low-frequency structural information \cite{QiZ2022CNNPostMultiScaleDWT}.
				Wang \textit{et al.} proposed a wavelet-based Generative Adversarial Network (GAN) to recover high-frequency details using a wavelet-domain generator \cite{WangJ2020MWGAN}.
				Recently, Kathariya \textit{et al.} fused DCT-domain features with spatial features or applied transformers to exploit long-range spectral correlations \cite{Kathariya2022MultiStageSpatial, Kathariya2022MultiStageLocally}.
				Zhang \textit{et al.} combined wavelet transform with PoolFormer to enhance spatial-spectral features \cite{Zhang2024ARF}.

		\subsection{Complexity-Oriented Design}	
			\subsubsection{Dynamic \& Early Exit}
				Xing \textit{et al.} introduced a Rate-Blind Quality Enhancement via an early-exit strategy, where easy samples exit the network at early layers while hard samples pass through the full depth, saving computation for simple regions \cite{XingQ2020EarlyExitQE}.
				Man \textit{et al.} proposed a dynamic in-loop filter where a policy network determines the optimal network topology for each block on-the-fly \cite{Man2025ContentAwareDI}.
			
			\subsubsection{Reconfigurable \& Scalable}
				Yang \textit{et al.} designed a scalable CNN where the number of active layers can be adjusted based on the available computational resources of the decoding device \cite{YangR2017ScalableCNNQE}.
				Zhang \textit{et al.} proposed a reconfigurable framework where the encoder searches for the optimal sub-network configuration and signals it to the decoder, balancing performance and complexity \cite{Zhang2024ARF}.
			
			\subsubsection{Lightweight \& Implementation Optimization}
				Song \textit{et al.} applied dynamic fixed-point arithmetic to address floating-point precision issues in hardware implementation \cite{SongX2018PracticalCNNILF}.
				Lee \textit{et al.} proposed a lightweight network estimating pixel-wise filter coefficients to bypass heavy CNN computations \cite{Lee2024PixelEF}.
				Lam \textit{et al.} \cite{Lam2020EfficientAdaptation} and Santamaria \textit{et al.} \cite{Santamaria2021ContentAdaptivePost} minimized overhead for adaptive filtering by fine-tuning and transmitting only the bias parameters of the model.
	
	\section{Performance and Complexity Analysis}\label{performance}
			Fig. \ref{fig:performance} shows the relationship between coding gain and complexity across three distinct dimensions: Complexity, Model Size, and Decoding Time. 
			In these scatter plots, the horizontal axis represents the respective complexity metric on a logarithmic scale to accommodate the wide range of varying architectures, while the vertical axis indicates the luma (Y) BD-rate reduction.
			Notably, the latest research trends demonstrate a paradigm shift towards ultra-lightweight architectures ($<50$k MACs/pixel) to satisfy the strict hardware constraints of emerging standards like JVET NNVC, proving that practical coding gains can be achieved without prohibitive complexity. 
			Based on this visual summary, this section provides a comprehensive review of performance relative to hardware feasibility.

		\subsection{Experimental Setup}
			In the case of datasets, DIV2K is the de-facto standard, providing 800 training and 100 validation high-resolution (2K) images. 
			Also, BVI-DVC is widely adopted, consisting of 800 sequences covering diverse motion and resolutions up to 4K. 
			Other utilized datasets include Vimeo-90K (containing 89,800 short clips) for temporal modeling and Common Test Condition (CTC) sequences (Class A-E) for validation.
			Regarding frameworks, the trend has shifted from MatConvNet and Caffe to PyTorch and TensorFlow. 
			Models are typically trained on cropped patches using MSE or L1 loss functions.

			The primary video coding standards targeted were HEVC/H.265 and its successor, VVC/H.266, evaluated using their respective reference software.
			Some studies also presented results for AV1 and older standards like H.264/AVC.
			They verified quantitative performance under the coding configuration defined by CTC.
			Performance was primarily measured by the BD-rate reduction and also used Peak Signal-to-Noise Ratio (PSNR).

		\subsection{Rate-Distortion Performance Comparison}
			\subsubsection{Evolution from Single to Multi-frame}
				Dai \textit{et al.} utilized PPF to achieve an average BD-rate reduction of 4.6\% compared to HEVC/H.265 intra coding \cite{DaiY2017CNNPostHEVCIntra}.
				Li \textit{et al.} improved upon this with residual learning, reaching approximately 5.5\% gain \cite{LiF2018DRNQE}.
				The explicit utilization of coding information proved crucial for further gains; He \textit{et al.} leveraged CU-maps to boost performance to 9.76\% \cite{HeX2018PartitionMaskQE}.
				Targeted at VVC/H.266, Huang \textit{et al.} achieved 11.1\% BD-rate reduction by incorporating prediction mode maps \cite{HuangH2021FrameWiseQE}.
				Multi-frame approaches exploit temporal correlations to surpass single-frame limits.
				Yang \textit{et al.} mitigated quality fluctuations \cite{YangR2018MultiFrameQE}.
				Ding \textit{et al.} achieved 9.00\% BD-rate reduction, comparable to the computationally expensive optical flow-based method, thus proving that temporal information can be utilized efficiently \cite{DingD2022BipredQE}.
				Other notable approaches include utilizing FlowNet2.0 for explicit motion compensation \cite{TongJ2019MultiFrameQE, MengX2020BSTN} or employing spatio-temporal fusion to handle complex motion \cite{ChenW2020NN_AR_Temporal}.
			
			\subsubsection{Depth, Adaptability, and Architecture}
				In ILF, Park \textit{et al.} replaced SAO with a shallow network, yielding modest gains of 1.9-2.8\% \cite{ParkWS2016CNNILF}.
				In contrast, deeper networks like Zhang \textit{et al.} \cite{ZhangY2018RHCNNILF} and Wang \textit{et al.} \cite{WangY2018DenseResidualILF} achieved significantly higher gains of 5.7\% (AI) and 6.9\% (AI), respectively, by employing residual highway units and dense connections.
				Moving to VVC/H.266, Huang \textit{et al.} utilized image gradients to achieve 3.29\% additional gain over VTM \cite{HuangZ2020MultiGradientILF}.
				To address the model redundancy issue, Huang \textit{et al.} employed a single model for multiple QPs and frame types, achieving 3.63\% (AI) and 3.56\% (RA) gains \cite{HuangZ2022OneForAll}.
				Ma \textit{et al.} reported the highest gains, exceeding 10\% in BD-rate, by employing a multi-level feature review mechanism, although this comes at the cost of high complexity \cite{MaD2021MFRNet}.
				Beyond CNNs, Transformer-based models have proven effective at capturing global dependencies. 
				For instance, the Swin Transformer-based filter achieved -7.05\% gain \cite{TongO2024Swin}. 
				However, recent standardization efforts suggest that simplifying Transformer-like structures by removing attention modules does not necessarily negatively impact BD-rate, suggesting potential for architectural optimization \cite{JVETZ0106}.
			
			\subsubsection{Ablation Studies on Coding Information}
				Ström \textit{et al.} conducted a rigorous ablation study on the NN-based ILF \cite{JVETZ0106}.
				They demonstrated that removing the explicit \textit{partitioning map} or \textit{boundary strength} input from the network resulted in negligible performance loss (maintaining a -7.57\% BD-rate in AI), while reducing computational complexity from 429 to 418 kMACs/sample. 
				In contrast, removing the \textit{prediction signal} caused a noticeable performance drop to -7.35\%.
				This indicates that prediction information is a critical feature for restoration, whereas explicit partitioning maps might be redundant for deep networks that can implicitly learn structural features from the pixel data.
	
		\subsection{Comparative Analysis by Filter Type and Coding Standard}
			\subsubsection{ILF vs. PPF}
				While Jia \textit{et al.} achieved gains of -4.1\% to -6.0\% using ILF, they increased the complexity of both the encoder and decoder \cite{JiaC2019ContentAwareILF}.
				In contrast, Ma \textit{et al.} only impacted the decoder side with PPF (-14.1\%) \cite{MaD2021MFRNet}.
				However, PPF still adds notable complexity to the decoder, which can be a bottleneck for real-time processing on CE devices.
			
			\subsubsection{Impact of Baseline Standards (HEVC/H.265 vs. VVC/H.266 vs. AV1)}
				For HEVC/H.265, top models achieve over -10\% BD-rate reduction.
				For VVC/H.266, gains are generally lower (3-5\% range for simple models) due to powerful built-in filters like ALF, though sophisticated models still achieve approximately -7.39\% (AI) gains \cite{JVETY0143}.
				Research on AV1 is also growing, with approaches replacing existing filters achieving -4.3\% to -3.2\% savings \cite{DingD2019CNNILFAV1}.
				Furthermore, Zou \textit{et al.} proposed a hybrid framework integrating an E2E Learned Image Codec (LIC) for intra frames with a conventional VTM inter-frame coder \cite{JVETAK0146}. 
				Although this approach demonstrated potential gains in RA, the decoding complexity remains a major bottleneck (e.g., 8.54 MMACs/pixel for the LIC decoder), highlighting the continued need for lightweight architectural innovations.
			
			\subsubsection{Fractional Interpolation Filter (FIF)}
				Pham \textit{et al.} achieved -4.5\% BD-rate for HEVC's RA by extending CNN interpolation to chroma components \cite{Pham2019DeepFI}.
				Notably, Ibrahim \textit{et al.} used a neural network to predict filter coefficients rather than replacing the filter directly, achieving up to -5.0\% gain with reduced complexity \cite{Ibrahim2018NNFME}.
				However, since FIF is invoked frequently within the ME/MC loop, the computational overhead per pixel is a critical constraint, often requiring extreme lightweighting or hardware acceleration.	

\begin{table}[t]
	\centering
	\caption{\scriptsize Standardization Feasibility Checklist: Key Engineering Constraints}
	\label{tab:std_checklist}
	\scriptsize
	\resizebox{\linewidth}{!}{
		\begin{tabular}{l|l|l|l}
			\hline\hline
			Category & 
			\begin{tabular}[c]{@{}l@{}}Constraint /\\ Metric\end{tabular} & 
			\begin{tabular}[c]{@{}l@{}}Target Requirement\\ (Standardization) \end{tabular}& 
			Engineering Implication \\\hline\hline
			\multirow{3}{*}{\begin{tabular}[c]{@{}l@{}}Implementation\\ \& Determinism\end{tabular}} & 
			\begin{tabular}[c]{@{}l@{}}Fixed-point\\ Support \end{tabular} & 
			\begin{tabular}[c]{@{}l@{}}Integer-only Inference (e.g., INT8/INT16)\\ *Floating-point operations are strictly prohibited.\end{tabular} & 
			\begin{tabular}[c]{@{}l@{}}Ensures bit-exact consistency across different\\ CPU/GPU/NPU platforms.\end{tabular} \\ 
			\cline{2-4} 
			& 
			\begin{tabular}[c]{@{}l@{}}Operation\\ Set \end{tabular} & 
			\begin{tabular}[c]{@{}l@{}}Limited Ops (Add, Shift, Clip, Mul)\\ *Avoid complex activations (e.g., Sigmoid, Tanh).\end{tabular} & 
			\begin{tabular}[c]{@{}l@{}}Reduces hardware logic gate count and latency;\\ Implementation for fixed-function hardware.\end{tabular} \\\hline
			\multirow{3}{*}{\begin{tabular}[c]{@{}l@{}}Computational\\ Complexity\end{tabular}} & 
			\begin{tabular}[c]{@{}l@{}}MACs /\\ Pixel \end{tabular} & 
			\begin{tabular}[c]{@{}l@{}}Bounded Complexity (e.g., $<$ 1k $\sim$ 2k MACs)\\ *Strict worst-case guarantee required.\end{tabular} & 
			\begin{tabular}[c]{@{}l@{}}Determines real-time decoding feasibility (e.g.,\\ 4K@60fps) within power budget.\end{tabular} \\ 
			\cline{2-4} 
			& 
			\begin{tabular}[c]{@{}l@{}}Parameter\\ Count \end{tabular} & 
			\begin{tabular}[c]{@{}l@{}}Compact Models (e.g., $<$ 64k parameters)\\ *Or shared weights across components.\end{tabular} & 
			\begin{tabular}[c]{@{}l@{}}Minimizes on-chip SRAM usage and external\\ DRAM bandwidth for weight fetching.\end{tabular} \\\hline
			\multirow{3}{*}{\begin{tabular}[c]{@{}l@{}}Memory\\ \& Bandwidth\end{tabular}} & 
			Line Buffer & \begin{tabular}[c]{@{}l@{}}Minimized Vertical Receptive Field\\ *Avoid full-frame buffering.\end{tabular} & 
			\begin{tabular}[c]{@{}l@{}}Critical for pipelined hardware architecture;\\ Affects chip die size directly.\end{tabular} \\ 
			\cline{2-4} 
			& 
			Granularity & 
			\begin{tabular}[c]{@{}l@{}}Block/CTU-level Processing\\ *Independence between blocks preferred.\end{tabular} & 
			\begin{tabular}[c]{@{}l@{}}Enables parallel processing and reduces\\ dependency-induced latency.\end{tabular} \\\hline
			Signaling & 
			Overhead & 
			\begin{tabular}[c]{@{}l@{}}Efficient Syntax (e.g., CABAC context)\\ *Avoid sending raw weights per slice.\end{tabular} & 
			\begin{tabular}[c]{@{}l@{}}Prevents coding gain dilution due to\\ heavy side-information transmission.\end{tabular} \\\hline\hline
		\end{tabular}
	}
\end{table}

		\subsection{Computational Complexity and Resource Efficiency}
			\subsubsection{Trade-offs between Model Size and Computation}
				Zhang \textit{et al.}'s model required approximately 3.34M parameters, doubling the encoding time \cite{ZhangY2018RHCNNILF}, whereas Zhang \textit{et al.} utilized recursive blocks to minimize parameters while maintaining depth \cite{ZhangS2020RRCNNILF}.
				Kang \textit{et al.} \cite{KangJ2017MultiModalILF} used 2.3M parameters, over 40 times that of Dai \textit{et al.} (54k) \cite{DaiY2017CNNPostHEVCIntra}.
				To provide a concrete example of high-performance complexity, the intra luma model \cite{JVETY0143} required approximately 429 kMACs/sample (649 kMACs/pixel for YUV) and 6.24M parameters to achieve a -7.39\% to -7.57\% gain \cite{JVETZ0106}.
				In contrast, the latest AHG report indicates that recent ILF candidates such as Low-complexity Operating Point 4 (LOP4) in NNVC-11.0 are constrained to approximately 1.2 kMAC/pixel and 36k parameters to achieve a 0.2\% coding gain in RA \cite{JVETAK0011}.
				This demonstrates a clear shift in the field from pure performance-driven designs to complexity-constrained optimization suitable for standardization.
			
			\subsubsection{Architectural Optimization for Hardware}
				To meet these stringent requirements, architectural optimizations are essential.
				Kim \textit{et al.} highlighted the effectiveness of Canonical Polyadic (CP) decomposition, which splits a standard $d \times d$ convolution into a sequence of lighter operations \cite{Kim2022Standardization}. 
				Theoretically, if the rank $R \cong \frac{ST}{S+T}$ (where $S, T$ are input/output channels), this technique reduces computational complexity from $STd^2$ to $(S+2d+T)R$, decreasing computation by a factor of approximately $d^2$.
			
			\subsubsection{Dynamic Complexity Control}
				Fixed computation for all inputs is inefficient.
				Xing \textit{et al.} introduced an early-exit mechanism, dynamically terminating inference for easy samples to reduce average computation \cite{XingQ2020EarlyExitQE}.
				Man \textit{et al.} utilized a policy network to dynamically generate optimal network topologies for each block, optimizing complexity without performance loss \cite{Man2025ContentAwareDI}.
				Yang \textit{et al.} proposed a scalable architecture where the network depth can be adjusted according to the device's computational capability \cite{YangR2017ScalableCNNQE}.
			
		\subsection{Hardware Implementation Considerations}
			\subsubsection{Precision and Platform Compatibility}
			Most models rely on 32-bit floating-point arithmetic, which incurs high hardware costs and drift issues.
			Song \textit{et al.} demonstrated that dynamic fixed-point arithmetic could minimize performance degradation while enabling integer-only operations, facilitating deployment on low-cost chips \cite{SongX2018PracticalCNNILF}.
			
			\subsubsection{Memory Bandwidth and Transmission Overhead}
			Multi-frame methods suffer from high memory bandwidth requirements.
			Lee \textit{et al.} proposed estimating pixel-wise filter coefficients via a lightweight network to bypass heavy CNN computations, reducing both memory access and computational load \cite{Lee2024PixelEF}.
			Furthermore, Lam \textit{et al.} and Santamaria \textit{et al.} proposed online adaptation techniques that fine-tune and transmit only the bias parameters, significantly reducing the update burden on the decoder and the bitstream overhead \cite{Lam2020EfficientAdaptation, Santamaria2021ContentAdaptivePost}.
			This approach makes DLF more feasible for bandwidth-constrained streaming environments.
			
	\section{Open Challenges and Future Directions}\label{challenges}
		While many academic works focus solely on RD performance, practical adoption in video coding standards requires strict adherence to engineering constraints. 
		To assist researchers in assessing the `standardization potential' of their methods, we present a Standardization Feasibility Checklist in Table \ref{tab:std_checklist}.
		This checklist distills the key requirements from recent standardization activities like JVET NNVC, covering four dimensions: implementation determinism, computational complexity, memory bandwidth, and signaling overhead.
		
		Recently, the DLF research paradigm has shifted from theoretical limits toward practical standardization and ultra-lightweight designs.
		Within JVET NNVC, efficient structures like cross-component reuse \cite{Li2025CCLOPCE} and spatial-channel mixing \cite{Cho2025SpatialChannelMB} were proposed to enhance performance while reducing computational complexity. 
		Additionally, domain-specific optimizations, such as utilizing auxiliary maps for SCC \cite{Li2025ScreenCV}, are expanding application scopes. 
		Finally, comprehensive tests evaluating the combination of multiple neural tools \cite{Qin2025CombinationTO} highlight ongoing efforts to bridge DL-driven efficiency with practical deployment constraints.

		\subsection{Computational Complexity and Hardware-friendly Design}
			The primary challenge is the prohibitive computational cost and memory bandwidth of deep models.
			Kang \textit{et al.} demonstrated the feasibility of efficient coding using multi-scale topologies \cite{KangJ2017MultiModalILF}. 
			Building on this foundation, Zhang \textit{et al.} and Zhao \textit{et al.} have further pushed the envelope by incorporating depth-wise separable convolutions and integer-only inference to meet the stricter constraints of modern hardware \cite{ZhangH2023Lightweight, Zhao2025AdvancedLC}.
			Kathariya \textit{et al.} \cite{Kathariya2022MultiStageSpatial} and Zhu \textit{et al.} \cite{Zhu2024NeuralNB} represented performance-oriented architectures to achieve significant coding gains. 
			However, their massive parameters pose severe challenges for practical deployment.
			While Song \textit{et al.} demonstrated the feasibility of dynamic fixed-point arithmetic \cite{SongX2018PracticalCNNILF}, future research must go beyond simple quantization.
			Techniques such as pixel-wise coefficient estimation by Lee \textit{et al.} \cite{Lee2024PixelEF} or dynamic topology generation by Man \textit{et al.} \cite{Man2025ContentAwareDI}, which adaptively adjust the computational path based on input complexity, are essential.
			In the case of Man \textit{et al.}, the policy network received the reconstructed frame and QP as input and decided whether to perform the filtering \cite{Man2025ContentAwareDI}. 
			
			Designing the architectures optimized for NPUs is critical for real-time video processing in CE devices. 
			NPUs are highly optimized for low-precision fixed-point arithmetic such as 8-bit/16-bit integer (INT8/INT16) and massive data parallelism. 
			Therefore, hardware-friendly DLFs should prioritize operations that minimize complex control logic and branching, which can stall the NPU's pipeline. 
			Furthermore, since memory bandwidth often becomes a bottleneck in 4K/8K video coding, architectures that support efficient data tiling and local memory caching are essential to minimize power-intensive off-chip memory access.
	
		\subsection{Generalization and Online Adaptability}
			DLFs trained on high-quality datasets often fail to generalize to heterogeneous content with different noise characteristics.
			The performance degradation in specialized content, such as SCC, can be attributed to a fundamental mismatch in statistical distributions.
			SCC consists of sharp edges, high-contrast text, and uniform flat areas, which exhibit different quantization noise characteristics compared to the natural textures found in common training datasets.
			Standard CNNs often fail to generalize to these `out-of-distribution' artifacts, as the learned filters are optimized for natural image priors.
			Although Huang \textit{et al.} improved adaptability to QPs and frame types \cite{HuangZ2022OneForAll}, its generalization to unseen resolutions or codec remains unproven.
			A promising solution lies in online fine-tuning, as proposed by Lam \textit{et al.} \cite{Lam2020EfficientAdaptation} and Santamaria \textit{et al.} \cite{Santamaria2021ContentAdaptivePost}, where only bias parameters are updated at the encoder and transmitted.
			Future work should explore meta-learning or few-shot adaptation techniques that allow the filter to adapt to new video domains with minimal data and updates, ensuring consistent performance across diverse content.

		\subsection{Temporal Consistency and Error Propagation Control}
			In PPF, independent frame processing often leads to flickering, as the lack of temporal coherence causes high-frequency fluctuations between frames \cite{SohJW2018DeepTemporalAR}. 
			Conversely, the primary concern is error propagation in ILF since filtered frames serve as reference pictures. 
			To address these issues, Ding \textit{et al.} employed a bi-directional prediction mechanism that aligns features from neighboring frames to enforce smoothness to suppress flickering in PPF \cite{DingD2022BipredQE}. 
			To mitigate error propagation in ILF, Lu \textit{et al.} leveraged Kalman filtering principles to track signal states over time, balancing noise reduction with the preservation of structural details required for prediction \cite{LuG2020DeepNonLocalKalman}. 
			Furthermore, with the advent of complex reference structures in VTM and ECM, state-based filtering that can explicitly track and block error propagation paths is required \cite{Zhao2025AdvancedLC}.

		\subsection{Perceptual Quality vs. Signal Fidelity}
			Generative methods like Wang \textit{et al.} \cite{WangJ2020MWGAN} and Li \textit{et al.} \cite{LiX2020MSGDN} excel in texture restoration but suffer from lower PSNR.
			This is critical in applications requiring high fidelity, such as medical imaging or broadcasting. 
			Building on the model selection signaling proposed by Nasiri \textit{et al.} \cite{Nasiri2021ModelSelectionQE}, future research should focus on controllable generative models.
			These would allow the coder to switch between modes based on the content characteristics or user requirements.	
			
	\section{Conclusion}\label{conclusion}
		This paper has presented a comprehensive survey and critical analysis of DLF for video coding standards, including HEVC/H.265 and VVC/H.266.
		We proposed a systematic taxonomy categorizing existing methods based on three pivotal dimensions: Integration Scheme within Video Coding, Coding Information Utilization, and Network Design Strategy.
		
		The key findings from this survey are summarized as follows: First, DLFs have demonstrated remarkable capabilities in mitigating complex non-linear artifacts that traditional hand-crafted filters (DBF, SAO, ALF) fail to address, thereby significantly improving both RD performance and visual quality. 
		In particular, approaches that actively exploit coding priors and temporal context have proven to be superior to simple pixel-based processing.
		
		Second, despite the performance gains, computational complexity and memory bandwidth remain the primary bottlenecks for commercial deployment. 
		While early academic studies understandably prioritized proving the theoretical upper bounds of coding efficiency, often resulting in prohibitive computational loads, this was a necessary proof-of-concept phase. 
		Building upon these foundational heavy models, the natural evolution of current research must now shift focus toward lightweight architectures, dynamic complexity control, and hardware-friendly designs.
		
		Third, standardization and compatibility are critical. 
		As evidenced by the JVET NNVC, for DL technologies to be adopted in future standards, they must satisfy stringent engineering requirements: specifically, integer-only fixed-point implementation via frameworks like SADL, strict caps on MACs/pixel, and the elimination of line-buffering to minimize on-chip memory usage.
		
		In conclusion, DLF is poised to become an indispensable component of next-generation video coding systems. 
		With the advancement of DL-specific accelerators (NPUs), we envision that intelligent video coding satisfying high performance, low power, and low latency will soon become a reality in IoT and CE.
		
	\section*{Acknowledgements}
	This work was supported by the Ministry of Science and Information Communication Technology (MSIT), Republic of Korea, under the Institute of Information \& Communications Technology Planning \& Evaluation (IITP) (RS-2025-25442867) and the National Research Foundation (NRF) (RS-2026-25489154) grants.

\bibliographystyle{IEEEtran}
\bibliography{references}
\vspace{-15px}
\begin{IEEEbiography}[{\includegraphics[width=1in,height=1.25in,clip,keepaspectratio]{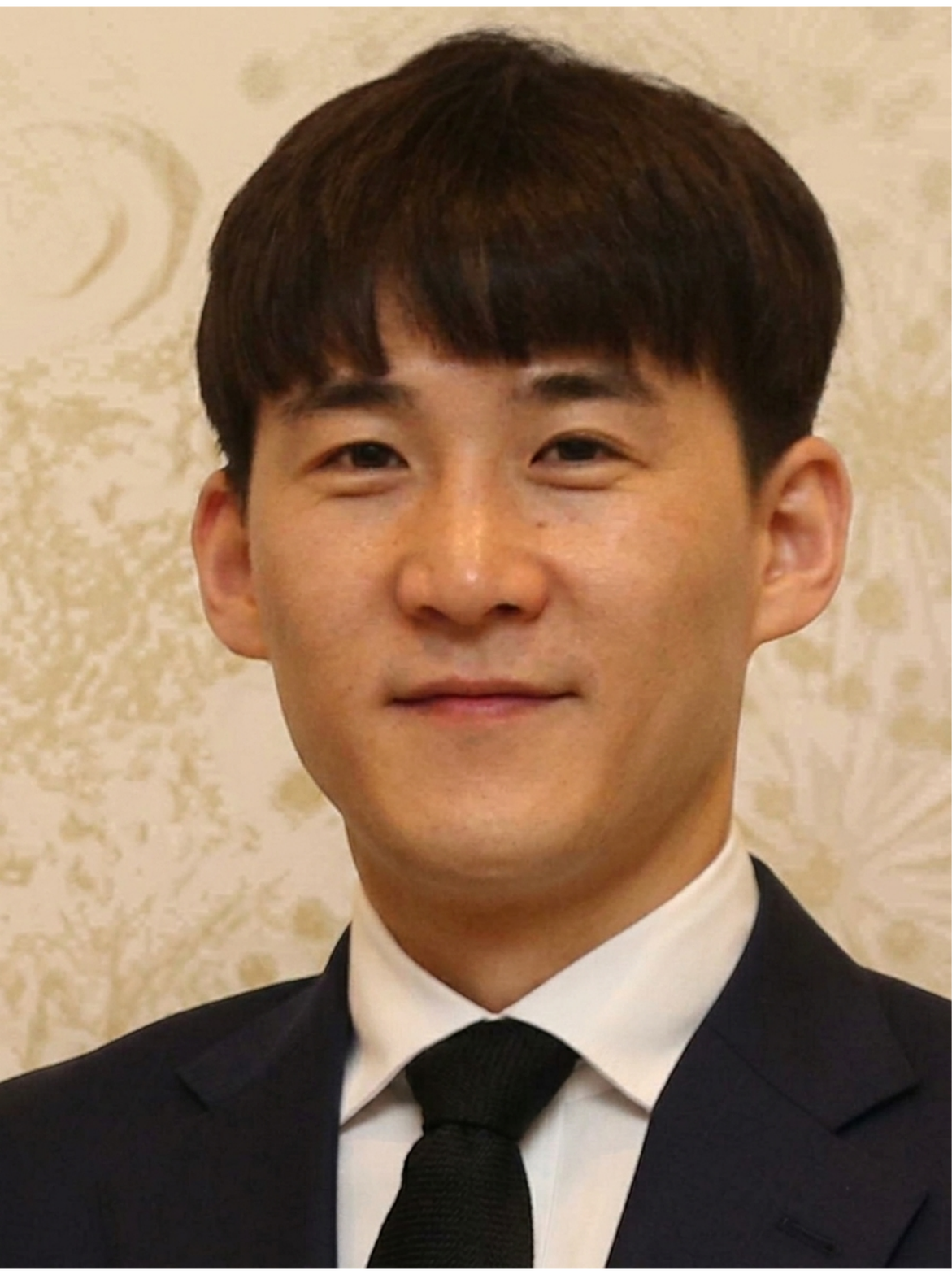}}]{Young-Woon Lee}
received the double B.S. degree
in public administration and computer engineer-
ing and the M.S. and Ph.D. degrees in computer
and electronics convergence engineering from Sun-
moon University, Asan, South Korea, in 2016, 2018,
and 2024, respectively. During his graduate studies,
his primary research focused on deep learning-
based video coding technologies, and he successfully
led numerous projects encompassing deep learn-
ing, computer vision, and image processing. Since
September 2025, he has been working as a Post-
Doctoral Researcher at the Electronics and Telecommunications Research
Institute (ETRI), Daejeon, South Korea. His current primary research focus
is the development of domain-specific vision-language models (VLMs). He
was honored with the Excellent Paper Award at the IEEE International
Conference on Consumer Electronics (IEEE ICCE) 2021, presented by the
IEEE Consumer Technology Society.
\end{IEEEbiography}
\vspace{-15px}
\begin{IEEEbiography}[{\includegraphics[width=1in,height=1.25in,clip,keepaspectratio]{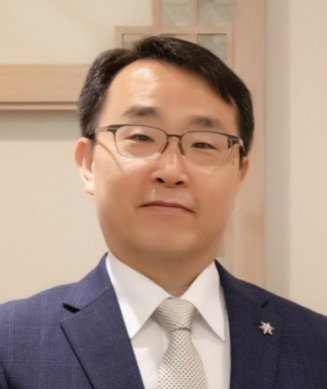}}]{Byung-Gyu Kim}
\textbf{(IEEE Senior Member)} received
the B.S. degree from Pusan National University,
Republic of Korea, in 1996, the M.S. degree from
Korea Advanced Institute of Science and Technol-
ogy (KAIST) in 1998, and the Ph.D. degree from
the Department of Electrical Engineering and Com-
puter Science, KAIST, in 2004. In March 2004, he
joined the Real-Time Multimedia Research Team,
Electronics and Telecommunications Research Insti-
tute (ETRI), South Korea, where he was a Senior
Researcher. In ETRI, he developed so many real-
time video signal processing algorithms and patents and received the Best
Paper Award in 2007. From February 2009 to 2016, he was an Associate
Professor with the Division of Computer Science and Engineering, Sun
Moon University, South Korea. In March 2016, he joined the Division of
Artificial Intelligence (AI) Engineering, Sookmyung Women’s University,
South Korea, where he is currently a Full Professor. He has published
over 300 international journal articles and conference papers, and patents in
his field. His major interests are computer vision algorithms, vision-based
deep learning models, multimodal generative AI mechanisms, and emotional
engineering. He is a Professional Member of ACM and IEICE. He also served
or serves on Organizing Committee Member of CSIP 2011, a Co-Organizer
of CICCAT2016/2017, The Seventh International Conference on Advanced
Computing, Networking, and Informatics (ICACNI 2019), and the EAI 13th
International Conference on Wireless Internet Communications Conference
(WiCON 2020), and the program committee members of many international
conferences. He has received the Special Merit Award for Outstanding Paper
from the IEEE Consumer Electronics Society, at IEEE ICCE 2012, the Certi-
fication Appreciation Award from the SPIE Optical Engineering, in 2013, and
the Best Academic Award from the CIS, in 2014. Also, he received Excellent
Paper Award, at IEEE International Conference on Consumer Electronics
(IEEE ICCE) 2021 (by IEEE CT Society). Also, he has been honored as the
World top 2
since 2022. He has been serving as a Professional Reviewer for many aca-
demic journals, including IEEE, ACM, Elsevier, Springer, Oxford, SPIE, IET,
MDPI, and IT\&T. He has been serving as an Associate Editor (AE) for IEEE
T RANSACTIONS ON C ONSUMER E LECTRONICS, IEEE ACCESS, Circuits,
Systems, and Signal Processing (Springer), The Journal of Supercomputing
(Springer), The Journal of Real Time Image Processing (Springer), Heliyon-
Computer Science (Cell Press), and Applied Sciences (MDPI). He has served
as the Editor-in-Chief (EiC) of Journal of Multimedia Information System
(JMIS) and Korea Multimedia Society (KCI indexed). For more information
\textcolor{blue}{\url{https://scholar.google.com/citations?user=Jl4y9tcAAAAJ\&hl}}
\end{IEEEbiography}

\end{document}